\documentclass[final,5p,times]{elsarticle}
\usepackage{graphicx}
\usepackage{subcaption}
\usepackage{epsfig}
\usepackage{siunitx}
\usepackage{booktabs}
\usepackage[T1]{fontenc}
\usepackage{multirow} 
\usepackage{tabularx}
\usepackage{array}
\usepackage[namelimits]{amsmath} 
\usepackage{amssymb} 
\usepackage{amsmath}            
\usepackage{amsfonts}            
\usepackage{mathrsfs} 
\usepackage{float}
\usepackage{cases}
\usepackage{wrapfig}
\usepackage{hyphenat}
\usepackage{multicol}
\usepackage{caption}
\usepackage[labelfont=bf, textfont=normalfont, justification=raggedright, singlelinecheck=false]{caption}
\usepackage{graphicx}
\usepackage{subcaption}
\usepackage{tabularx}  
\usepackage{amsmath}

\usepackage{amssymb}

\journal{arxiv}
\date{}

\begin{document}
\begin{frontmatter}



\title{ A Robust Error-Resistant View Selection Method for 3D Reconstruction \tnoteref{13}}


\tnotetext[13]{This work was supported in part by the National Natural Science Foundation of China (No. 62172190), National Key Research and Development Program(No. 2023YFC3805901), the "Double Creation" Plan of Jiangsu Province (Certificate: JSSCRC2021532) and the "Taihu Talent-Innovative Leading Talent" Plan of Wuxi City(Certificate Date: 202110). }
\author[label1]{Shaojie Zhang}
\ead{7213107006@stu.jiangnan.edu.cn}
\author[label1,label2]{Yinghui Wang\corref{cor1}}
\ead{wangyh@jiangnan.edu.cn}
\author[label1]{Bin Nan}
\ead{binnan98@qq.com}
\author[label1]{Wei Li}
\ead{cs_weili@jiangnan.edu.cn}
\author[label1]{Jinlong Yang}
\ead{yjlgedeng@163.com}
\author[label1]{Tao Yan}
\ead{yantao.ustc@gmail.com}
\author[label1]{Yukai Wang}
\ead{ericwangyk22@163.com}
\author[label3]{Liangyi Huang}
\ead{lhuan139@asu.edu}
\author[label4]{Mingfeng Wang}
\ead{mingfeng.wang@brunel.ac.uk}
\author[label5]{Ibragim R. Atadjanov}
\ead{ibragim.atadjanov@gmail.com}
\cortext[cor1]{Corresponding author}
\affiliation[label1]{organization={ School of Artificial Intelligence and Computer Science, Jiangnan University},
            addressline={1800 Li Lake Avenue},
            city={wuxi},
            postcode={214122},
            state={Jiangsu},
            country={PR China}}
\affiliation[label2]{organization={ Engineering Research Center of Intelligent Technology for Healthcare, Ministry of Education},
            addressline={1800 Li Lake Avenue},
            city={wuxi},
            postcode={214122},
            state={Jiangsu},
            country={PR China}}
 \affiliation[label3]{organization={School of Computing and Augmented Intelligence, Arizona State University},
            addressline={1151 S Forest Ave},
            city={Tempe},
            postcode={8528},
            state={AZ},
            country={U.S}}
\affiliation[label4]{organization={Department of Mechanical and Aerospace Engineering, Brunel University},
            addressline={Kingston Lane},
            city={London},
            postcode={UB8 3PH},
            state={Middlesex},
            country={U.K}}    
\affiliation[label5]{organization={Tashkent University of Information Technologies named after al-Khwarizmi},
			addressline={ 108 Amir Temur Avenue},
			city={Tashkent},
			postcode={100084},
			state={},
			 country={Uzbekistan}}

\begin{abstract}
To address the issue of increased triangulation uncertainty caused by selecting views with small camera baselines in Structure from Motion (SFM) view selection, this paper proposes a robust error-resistant view selection method. The method utilizes a triangulation-based computation to obtain an error-resistant model, which is then used to construct an error-resistant matrix. The sorting results of each row in the error-resistant matrix determine the candidate view set for each view. By traversing the candidate view sets of all views and completing the missing views based on the error-resistant matrix, the integrity of 3D reconstruction is ensured. Experimental comparisons between this method and the exhaustive method with the highest accuracy in the COLMAP program are conducted in terms of average reprojection error and absolute trajectory error in the reconstruction results. The proposed method demonstrates an average reduction of 29.40\% in reprojection error accuracy and 5.07\% in absolute trajectory error on the TUM dataset and DTU dataset.\end{abstract}



\end{frontmatter}


\section{Introduction}
View selection refers to determining the degree of correlation among images in Structure-from-Motion (SfM) methods. It involves selecting correlated views for matching in the next reconstruction, thus determining the reconstruction order. This is a critical technique with a dual impact on accuracy and speed in multi-view 3D reconstruction, making it an indispensable strategy in SfM.

In 3D reconstruction, the choice of views with different camera baselines leads to variations in triangulation errors, subsequently affecting the accuracy of 3D point recovery during the triangulation phase. The influence of camera baseline on triangulation error can be characterized as follows: the range of triangulation error sharply decreases with an increase in baseline, reaching a minimum point, after which the error range grows slowly. However, classical SfM methods \cite{ref1} tend to select views with smaller camera baselines based on content similarity or feature matching results during view selection, resulting in an increase in triangulation error range and a subsequent decrease in 3D reconstruction accuracy.

To avoid selecting views with smaller camera baselines during view selection, some SfM methods \cite{ref2}-\cite{ref3} set a threshold for the camera baseline. Only views that meet the threshold can participate in reconstruction. However, since the minimum error points vary for different datasets in the influence of camera baseline on triangulation error, fixed thresholds cannot effectively handle diverse reconstruction environments. Therefore, setting a threshold does not fundamentally address the problem of selecting views with smaller baseline errors. Currently, a better solution is graph-based methods \cite{ref4}-\cite{ref5}, which optimize a cost function related to camera baseline to evaluate the final reconstruction quality. This approach determines the relationships among views and the reconstruction order. However, this method requires computing the reconstruction accuracy for the remaining views each time a view for the next reconstruction is selected, significantly increasing the computational complexity of the view selection process, particularly reducing operational efficiency. Since the camera baseline accurately reflects triangulation errors and impacts 3D reconstruction accuracy, guiding the view selection process based on this criterion can significantly improve 3D reconstruction accuracy while ensuring efficiency. Therefore, this paper proposes an error-resistant view selection method, with main contributions and advantages summarized as follows:

(1) A view selection method based on camera baseline with triangulation error resistance as an indicator is proposed. This method can address the issue of increased triangulation uncertainty resulting from choosing a smaller camera baseline.

(2) A recursive strategy for view omission and completion is designed to ensure the participation of all views in 3D reconstruction, thereby enhancing the accuracy of the reconstruction.

(3) The integration of the proposed method with the COLMAP application is implemented, and experiments are conducted on DTU and TUM datasets. The results validate that the view selection method proposed in this paper, relative to the exhaustive approach of COLM\\AP, achieves a 29.40\% improvement in terms of reprojection error.

\begin{figure*}[t]
    \centering
    \includegraphics[height=3.6in]{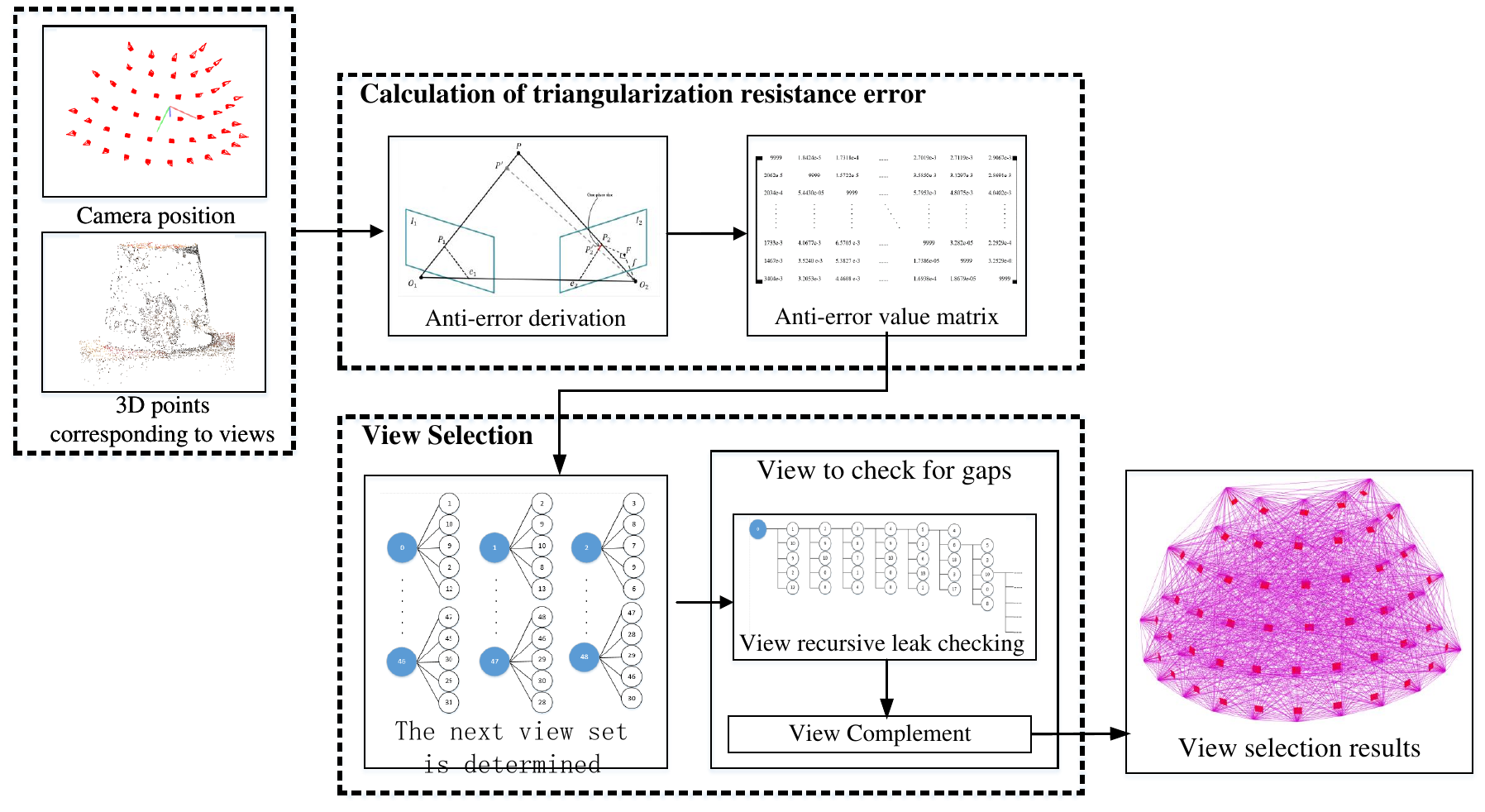}
\caption{Methodological Framework.}
    \label{Fig1}
\end{figure*}

\section{RELATED WORK}
Early Structure-from-Motion (SfM) methods \cite{ref2},\cite{ref6}-\cite{ref9} often employed exhaustive search to determine candidate images for matching, and the order of view reconstruction was based on the number of feature matches. While such methods ensured the availability of SfM on ordered or unordered image sets with high precision, they led to severe computational inefficiencies for larger image datasets. Subsequent improvements categorized view selection methods into two types based on the similarity of either view content or scene geometry. Approaches based on view content similarity emphasize reconstruction efficiency, whereas those based on scene geometry similarity focus on reconstruction accuracy.

To significantly enhance the efficiency of view selection, many scholars have utilized the bag-of-words model \cite{ref12} to determine relevant views based on content similarity and imposed limits on the number of candidate views. For instance, Nister et al. [13] proposed an efficient query for the most similar candidate views using a vocabulary tree implemented with the bag-of-words model and hierarchical clustering. Subsequently, Agarwal et al. \cite{ref14} introduced a view selection method based on the bag-of-words model and random projection trees, where each view only matches the 10 most visually similar views, reducing the number of matches. Jiang et al. \cite{ref15} achieved an efficient and reconstruction-accurate view selection method for large-scale drone image sets using the index structure of a vocabulary tree. Siedlaczek et al. \cite{ref16} optimized the top-k query algorithm in a bag-of-words model-based view selection method, greatly improving the speed of view selection. These methods, based on content similarity, are advantageous for improving the efficiency of determining candidate views, and content similarity is advantageous for image matching. However, views associated based on content similarity generally have smaller baseline lengths, which is detrimental to the accuracy of the triangulation phase.

Methods based on scene geometry similarity for view selection are mainly implemented through clustering and graph theory. For example, Li et al. \cite{ref11} determined a set of landmark views based on image descriptors and geometric constraints, constructing a graph structure by associating camera poses for landmark views in each set. During reconstruction, priority is given to selecting views with the shortest camera distances. This method, by selecting smaller camera baselines, is beneficial for reducing errors in the image matching step but overlooks that smaller camera baselines can amplify errors in the triangulation phase, thus hindering the improvement of reconstruction accuracy. To avoid improper baseline usage, Snavely et al. \cite{ref10} treated images as skeleton nodes, using two-view reconstruction to recover camera poses and 3D points for the skeleton views. They considered the reconstruction accuracy as skeleton edges and selected skeleton subgraphs that satisfied covariance limits for incremental reconstruction. This effectively ensured that the choice of camera baseline promoted the improvement of reconstruction accuracy but still relied on exhaustive methods, significantly increasing computational time.

Therefore, choosing appropriate camera baselines to enhance 3D reconstruction accuracy while ensuring the efficiency of view selection remains a problem that has not been fully addressed. This paper proposes an error-resistant view selection method that quantifies the error resistance of camera baselines to triangulation. This method avoids the complex iterative process of solving for all baselines in existing methods, efficiently selecting more reasonable camera baselines, and ensuring an improvement in 3D reconstruction accuracy.

\section{METHODOLOGY}
The technical framework of the view selection method based on camera baseline in this paper is illustrated in Figure 1. It consists of two main components: triangulation error resistance calculation and view selection.

In the triangulation error resistance calculation, the focus is on computing the triangulation error caused by a 1-pixel matching error under different camera baselines. This error value is considered as an indication of the baseline's resistance to errors in the triangulation phase. The errors for all pairwise baselines formed by corresponding views are calculated, creating a matrix of error resistance values.

Subsequently, in the view selection component, each row of the error resistance matrix is sorted in ascending order. The original indices of the top five elements in each row are selected, yielding the candidate set for the next reconstructed view for each image. Following the selection process of the next reconstructed view in the SfM workflow, with error resistance values as the criterion, recursive evaluation is performed for the candidate views corresponding to each image. Recursively marked views are identified, and the remaining unmarked views are considered as missing views. By adding the missing views to the candidate set corresponding to the baseline view with the minimum error resistance value, the process of identifying omitted views is completed. The next view set for all views, after the completion, represents the result of view selection.

\subsection{Triangulation Error Resistance Calculation}
\subsubsection{Camera Baseline Error Resistance Mechanism}
In the triangulation process, the specific coordinates of a 3D point can be reconstructed by utilizing the pixel points on the imaging planes of two cameras, along with the camera baseline. However, the corresponding pixel points determined through feature matching between two views contain errors. Moreover, the triangulation error resulting from the same feature matching error varies under different camera baselines. Therefore, different camera baselines exhibit error-resistant characteristics during triangulation, as illustrated in Figure 2(a). Here, t represents the distance of the camera baseline, $\delta \theta$  is the angle between the estimated projection line of the 3D point and the actual projection line, and  $\delta \theta$ arises due to errors in feature matching. $\delta d$ represents the triangulation error, indicating that the same $\delta \theta$  error will result in a larger triangulation error $\delta d$ for a smaller camera baseline $t$.

To illustrate the pattern of triangulation errors caused by the same feature matching error under different camera baselines, fixed coordinates of point $P$ and a fixed angle $\delta \theta $ are assumed. The triangulation error $\delta d$ is then calculated for different camera baselines $t$, and a curve depicting the relationship between baseline and triangulation error is plotted, as shown in Figure 2(b). It can be observed that the triangulation error sharply decreases as the baseline increases, reaching a minimum point, after which the error increases slowly. While this trend is generally applicable, the gradient and the location of the minimum point for each point on the curve need to be calculated based on specific numerical values.
\begin{figure}[H]	
	\centering
	\begin{subfigure}{0.45\linewidth}
		\centering
		\includegraphics[width=0.95\linewidth]{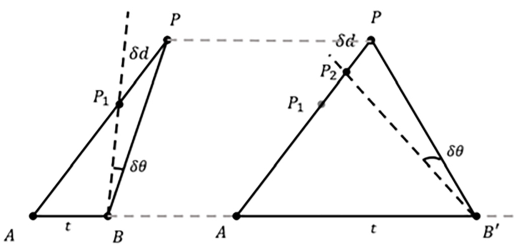}
		\caption{Graph of the effect of baseline on triangulation error}
		\label{Fig21}%
	\end{subfigure}
	\centering
	\begin{subfigure}{0.45\linewidth}
		\centering
		\includegraphics[width=0.95\linewidth]{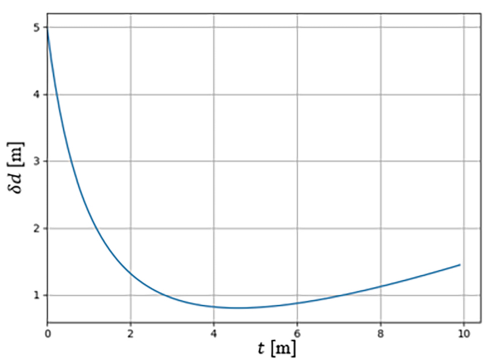}
		\caption{Regularity of the effect of baseline on triangulation error}
		\label{Fig22}
	\end{subfigure}
	\caption{Relationship between baseline and triangulation.}
\end{figure}
\subsubsection{Calculation of Error Resistance Values}
The specific expression of error resistance for different camera baselines in the triangulation phase is as follows: by assuming that the error in feature matching results in a 1-pixel error on the corresponding epipolar line, the error in solving the 3D point due to this error is calculated. This error value is referred to as triangulation error resistance value, and a smaller value indicates a stronger error resistance for triangulation under that baseline.

The schematic diagram for the derivation of triangulation error resistance values is shown in Figure 3. In Figure 3(a), $O_{1}$ and $O_{2}$ represent two cameras, $I_{1}$ and $I_{2}$ are the physical imaging planes corresponding to cameras $O_{1}$ and $O_{2}$, $P$ is the 3D point recovered based on the actual camera poses, $P_{1}$ and $P_{2}$ are the projection points of 3D point $P$ on physical imaging planes $I_{1}$ and $I_{2}$, $e_{1}$ and $e_{2}$ are epipolar points, and $f$ is the focal length. The calculation involves introducing a 1-pixel error on $P_{2}$ along its epipolar line $P_{2}e_{2}$ due to an error in feature matching, resulting in the triangulation error of the reconstructed 3D point ${P}'$, denoted as ${PP}'$. This error is considered as the triangulation error resistance value for the baseline $\vec{   O_{1}O_{2}   }$. It is important to note that the triangulation error resistance value for the baseline vector $\vec{   O_{1}O_{2}   }$ is not equivalent to the value for the baseline vector $\vec{   O_{1}O_{2}   }$.
\begin{figure}[H]	
	\centering
	\begin{subfigure}{0.45\linewidth}
		\centering
		\includegraphics[width=0.95\linewidth]{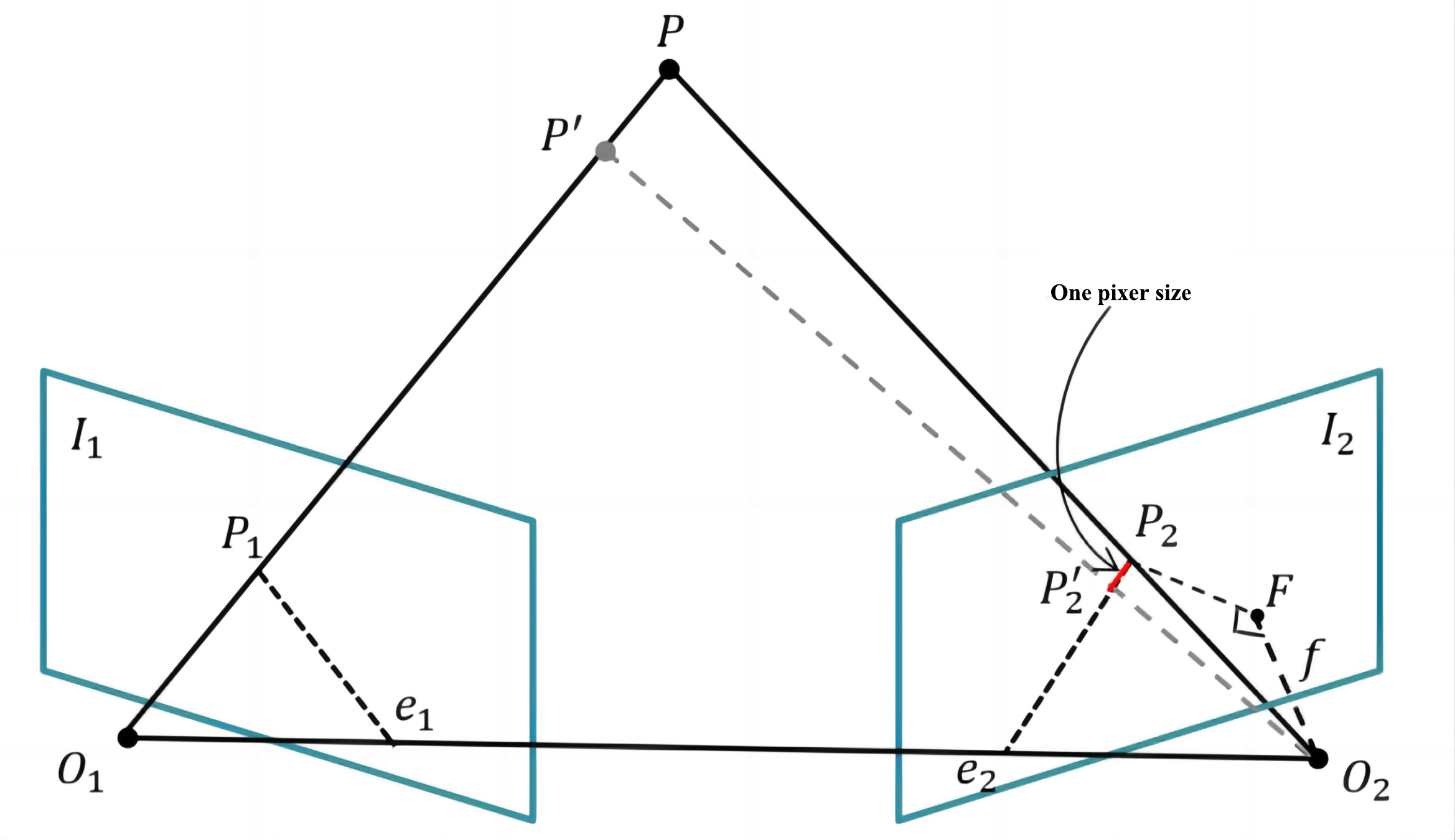}
		\caption{Triangulation Model}
		\label{Fig31}%
	\end{subfigure}
	\centering
	\begin{subfigure}{0.45\linewidth}
		\centering
		\includegraphics[width=0.95\linewidth]{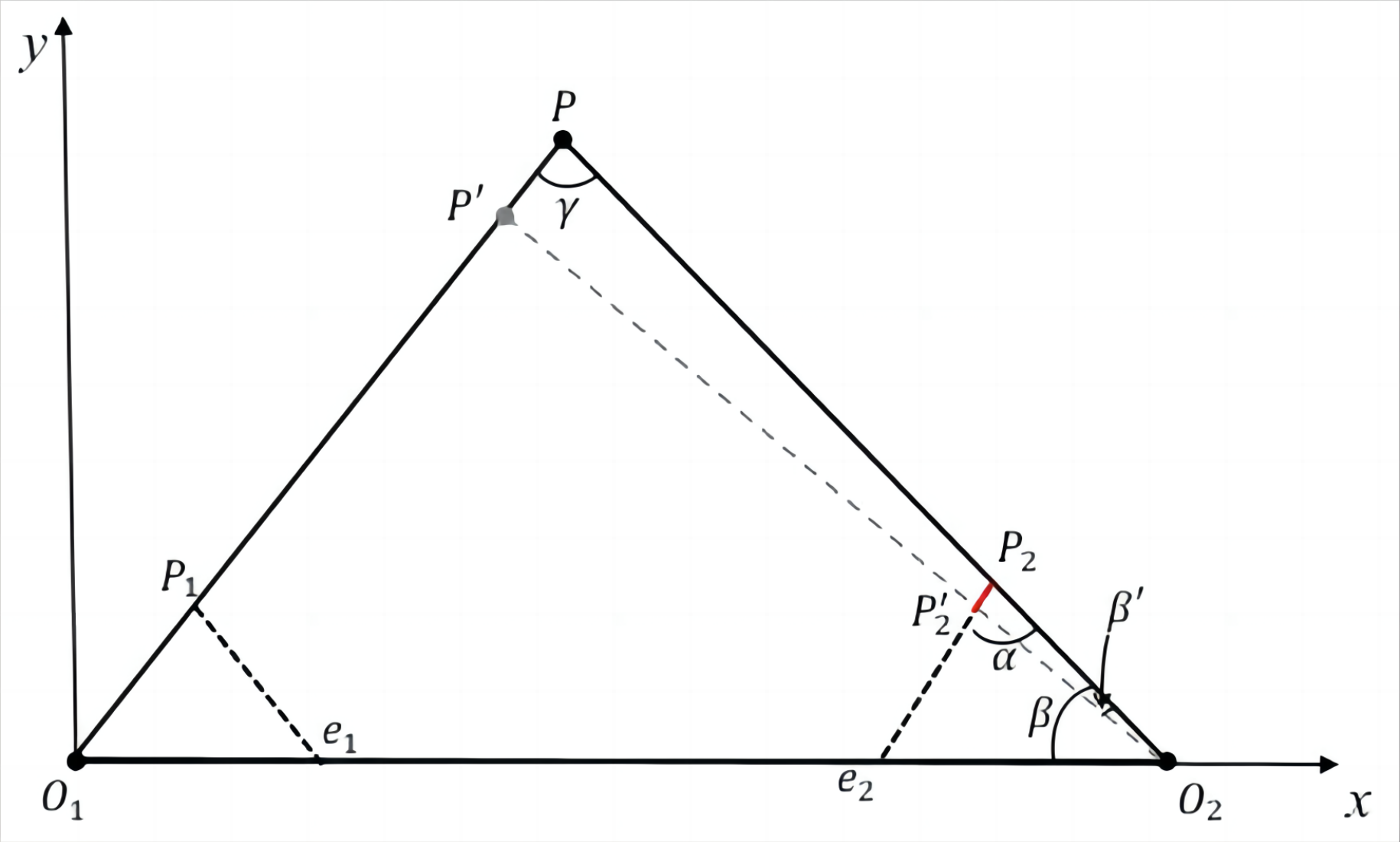}
		\caption{Triangulation Error Plan View}
		\label{Fig32}
	\end{subfigure}
	\caption{Schematic Diagram of Triangulation Error Resistance Principle.}
\end{figure}

Given the coordinates of points  $O_{1},  O_{2}$, and P in world coordinates, as well as the direction of the vector $\vec{   O_{2}F   }$, where the focal length f is in pixel units and its actual size is related to real-world scale, with the proportional impact on all segment sizes, and since any chosen value can be considered without affecting the overall result, we set the actual size of $O_{2}F$ to be 1. Using the known coordinates of the points and applying the law of cosines, the following angles can be determined: $\angle PO_{2}F$, $\angle O_{1}O_{2}F$,and $\angle O_{1}PO_{2}$. Subsequent calculations yield the lengths of the corresponding line segments, as shown in Equation (1).
\begin{equation}
	\label{eq1}
	\left\{\begin{matrix}
		O_{2}P_{2}=1/cos\angle PO_{2}F\\ 
		O_{2}e_{2}=1/cos\angle O_{1}O_{2}F
		\\e_{2}P_{2}=\sqrt{O_{2}P^{2}+O_{2}e_{2}^{2}-2O_{2}e_{2}\cdot O_{2}e_{2}\cdot cos\angle PO_{2}O_{1} } 
		\end{matrix}\right.      
\end{equation}

Additionally, for ease of expression, as depicted in Figure 3(b), a 2D coordinate system is established in the plane containing $ O_{2}$, $ O_{1}$, and $P$.  $O_{1}$ is taken as the origin, the direction of the vector $\vec{   O_{1}O_{2}   }$ is considered the positive x-axis, and the direction towards point P, perpendicular to $\vec{   O_{1}O_{2}   }$ and passing through $ O_{1}$, is designated as the positive y-axis. This establishes a 2D Cartesian coordinate system in the $O_{2}O_{1}P$ plane. Based on Figure 3(b) and Equation (1), the triangulation error resistance value ${PP}'$ for the baseline $\vec{   O_{1}O_{2}   }$ can be determined, as shown in Equation (2).

\begin{equation}
	\label{eq2}
	\left\{\begin{matrix}
		{PP}'=\frac{O_{2}P}{sin(\gamma + {\beta}')}sin{\beta }'\\ 
		sin{\beta }'=\frac{{P_{2}}'P_{2}}{{P_{2}}'O_{2}}sin\alpha \\ 
		{P_{2}}'O_{2}=\sqrt{O_{2}P_{2}^{2}+{P_{2}}'P_{2}^{2}-2O_{2}P_{2}\cdot{P_{2}}'P_{2}\cdot cos\alpha  }\\ 
		sin\alpha = \frac{e_{2}O_{2}}{e_{2}P_{2}}sin\beta 
		\end{matrix}\right.   		
\end{equation}
Where, \(O_2P\), \(O_2P_2\), \(O_2e_2\), and \(e_2P_2\) are known values obtained from Equation (1), \(\gamma\) is the angle \(\angle O_1PO_2\), \(\alpha\) is the angle \(\angle e_2P_2O_2\), \(\beta\) is the angle \(\angle PO_2O_1\), and \(\beta'\) is the angle \(\angle PO_2P'\).
\subsubsection{Construction of Error Resistance Matrix}
Building upon the derivation of error resistance values as described above, the corresponding error resistance values for the $n(n-1)$ baseline vectors formed by the existing $n$views are calculated. It is important to note that in the context of this paper, baselines are directional due to the different reconstruction orders of views from cameras $O_{1}$ and $O_{2}$ during the 3D reconstruction phase. As illustrated in the triangulation model diagram in Figure 2(a), the error resistance value for the baseline $\vec{   O_{1}O_{2}   }$ is along the segment $O_{1}P$, while the error resistance value for the baseline $\vec{   O_{1}O_{2}   }$ is along the segment $O_{2}P$. This indicates that the error resistance values corresponding to the baselines $\vec{   O_{1}O_{2}   }$and $\vec{   O_{2}O_{1}   }$ are different.

To facilitate subsequent view selection, the error resistance values for the $n(n-1)$ baseline vectors are integrated into a matrix denoted as E. The matrix E is a square matrix of size n, where the value E[i][j] represents the error resistance value for the baseline from camera i to camera j. The diagonal elements of matrix E correspond to the baselines from camera i to itself, which are meaningless. To prevent interference with subsequent calculations, these diagonal values are set to the maximum value of the variable type. 
\subsection{View selection}
\subsubsection{Determination of Next View Set}
For ease of expression, the candidate set for the next reconstructed view for each view is referred to as the "next view set." The next view set for view $S_i$ is determined based on the error resistance matrix. This set comprises the 5 views with the minimum error resistance values when viewed from the perspective of view $S\_i$ as the starting point.

Specifically, when determining the next view set for view $S\_i$, the values in the i-th row of the error resistance matrix E are selected and sorted in ascending order. The original indices of the top five values after sorting represent the indices of the candidate views.

Mathematically, this process can be expressed as follows: Let $E_i$ be the set of error resistance values for the baselines formed between view $S_i$ and all other views, i.e., $E_{i}=\begin{Bmatrix} E_{i1},\cdots  \end{Bmatrix} $. Here, $E_{ij}$ represents the error resistance value between view $S_i$ and view $S_j$. Sorting the elements in set $E_i$ in ascending order yields $E_{i}=\begin{Bmatrix} E_{ia_{1}},E_{ia_{2}},\cdots ,E_{ia_{n}}\mid E_{ia_{1}}< E_{ia_{2}}< \cdots < E_{ia_{n}} \end{Bmatrix}$. Therefore, the next view set for view $i$ is $\begin{Bmatrix}
	S_{a_{1}},S_{a_{2}},S_{a_{3}},S_{a_{4}},S_{a_{5}}
	\end{Bmatrix}$.
\subsubsection{View Check and Completion}
Due to the efficiency considerations in the view selection method proposed in this paper, each view's next view set consists of only 5 views. It is highly possible that some views may not be present in the next view set of all views and thus are not included in the 3D reconstruction process. To facilitate the understanding of this step, mathematical symbols are used for description. Let there be n views, denoted as the set $S=\begin{Bmatrix} S_{i}\mid1,2,\cdots n \end{Bmatrix}$. 
The next view set for view $S_i$ is represented as $C_{i}=\begin{Bmatrix} C_{ij}\mid j=1,2,3,4,5,C_{ij}\epsilon S \end{Bmatrix}$. To verify if all views are covered in the next view sets, it is checked whether $S=C_{1}\cup  C_{2}\cup \cdots \cup C_{n}$.

Based on the recursive process of selecting the next view in SfM reconstruction, combined with the error-resistant view selection method proposed in this paper, the process of checking and completing views is as follows: Select the baseline with the minimum error resistance value among all baselines. Consider the view corresponding to the starting point of this baseline as the initial view and begin recursion from this view. Mark the current view and choose the view with the minimum error resistance value from the next view set of the current view as the next view. During this process, two special cases may occur:

(1) If the next view is already marked, choose the second-best view from the next view set of the current view.

(2) If all views in the next view set of the current view are marked, return to the previous iteration's current view. Choose an unmarked view from the next view set at this point as the next view.

Continue iterating with the next view as the current view until the current view returns to the initial view, and all views in its next view set are marked. At this point, the recursion ends. The unmarked views are those that have not participated in the reconstruction. For each unreconstructed view $S_x$, find the baseline with the minimum error resistance value that terminates at that view $\vec{   S_{a}S_{x}   }$, and add view $S_x$ to the next view set of view  $S_{a}$. 
\section{Experimental and Result Analysis}
\subsection{Evaluation Metrics}
As view selection is just one component of the SfM method, its evaluation is ultimately based on the 3D scene reconstructed by SfM. This paper evaluates the view selection method from the perspectives of the accuracy of recovered 3D points and the precision of camera pose estimation. The assessment involves using the average reprojection error to evaluate 3D point accuracy and the absolute trajectory error to evaluate camera pose estimation accuracy.
\subsubsection{Average Reprojection Error}
Reprojection error measures the distance between the projection of a 3D point in camera coordinates and the actual pixel on the image, considering homography matrices and triangulation calculations. Using the average reprojection error provides a better reflection of the accuracy of 3D points recovered by the SfM method. The calculation of the average reprojection error is shown in Equation (3).
\begin{equation}
	\label{eq3}
	E= \frac{1}{c}\sum_{i=1}^{m}\sum_{j=1}^{n}\left \|x_{ij}-\pi(P_{i}X_{j}) \right \|\begin{matrix}
		2\\ 2
		\end{matrix} 		
\end{equation}
Where $m$ is the number of views. $n$ is the number of 3D points. $c$ corresponds to the number of 2D points,p is the projection matrix by which the camera projects the scene onto view $i$, $x_{j}$ is a 3D point, and $x_{ij}$ is the 2D coordinate of a 3D point in view $i$. $\pi((x,y,z)^{T})= (x/z,y/z)^{T}$ is the projection function, used to convert homogeneous coordinates into 2D coordinates
\subsubsection{Absolute Trajectory Error}
Absolute trajectory error calculates the difference between the estimated and true camera poses, reflecting the accuracy of camera pose estimation in the SfM method. As the world coordinate system established by the SfM method differs from the coordinate system corresponding to the true poses, alignment between the two is necessary. The method proposed by Zhang et al. \cite{ref17} for multi-state alignment is used. The core of this method is to solve for a similarity transformation matrix satisfying Equation (4), minimizing the Euclidean distance between the estimated and true trajectories.
\begin{equation}
	\label{eq4}
	{S}'= argmin\sum_{i=0}^{N-1}\left \|P_{i}-sR({P_{i}}'-t) \right \|^{2}		
\end{equation}
Where: $p_{i}$ is the true coordinates of the camera position $i$.
- ${P_{i}}'$ is the estimated coordinates of the camera position $i$.
- $s$ is the scaling parameter in the similarity transformation matrix.
- $R$ is the rotation matrix in the similarity transformation matrix. 
- $t$ is the translation matrix in the similarity transformation matrix. 
\subsection{Dataset}
This study primarily utilizes two types of datasets: the TUM dataset for indoor scenes \cite{ref18} and the DTU dataset for object models \cite{ref19}. The TUM dataset consists of video sequences captured by a handheld RGB-D camera in slow motion, depicting office scenes. For this experiment, two video sequences were selected: "freiburg1\_xyz," showcasing an office desk, and "freiburg1\_teddy," featuring an indoor teddy bear scene. The DTU dataset comprises images captured by an industrial robot at 49 or 64 camera positions for various object models. This study uses 20 image sets from the DTU dataset for experimentation.

To validate the effectiveness of the proposed view selection method across image sets with different baselines, the TUM dataset was processed. Different baseline image sets were obtained by selecting images with varying frame intervals. The frame intervals were set from 0 to 6 frames.
\subsection{Parameter Determination Experiment}
The main parameter in this study is the number of views in each view's next reconstruction set. If the set contains too few views, such as only one, the reconstruction stage is prone to failure due to either not meeting the registration requirements for the 3D reconstruction view or excessively large reprojection errors of the reconstructed 3D points. This can result in the exclusion of that view with no other candidate views, leading to the failure of the entire 3D reconstruction. On the other hand, if the number of candidate views is increased, it will result in more image matching iterations. Additionally, during the 3D reconstruction when selecting the next reconstruction image, more candidate images need to be traversed. Therefore, the number of views in the next view set should strike a balance between the feasibility and efficiency of 3D reconstruction.

It is worth noting that, due to the iterative nature of 3D reconstruction, setting different numbers of views in the set will influence the order of view selection. Although this method prioritizes the anti-error values of camera baselines and determines a fixed order for view reconstruction, the results might not meet the specified conditions, such as non-convergence in pose estimation or insufficient percentage of 3D points satisfying constraints. In such cases, views will be discarded and new views will be selected for iteration.

In this parameter experiment, the study tested the reconstruction time and reprojection error under different numbers of candidate views on datasets with 200 and 500 consecutive images selected from the "freiburg1\_teddy" sequence. The experiment aimed to explore the impact of the number of candidate views on reconstruction time and accuracy on datasets of varying scales. The results are presented in Table 1. In Table 1, the symbol "$\times$" indicates reconstruction failure. While this section only conducted experiments on two datasets, it can be inferred simplistically, based on the understanding of 3D reconstruction, that as the number of images in the 3D reconstruction set increases, i.e., the dataset scale becomes larger, more candidate views are needed to ensure the completion of 3D reconstruction. The table reflects the trend that as the number of candidate views increases, the reconstruction time significantly increases. There is no clear pattern in the impact of the number of candidate views on the reprojection error, but it shows a noticeable decrease when the number of candidate views exceeds 4 or 5 on the two datasets. Therefore, to ensure the feasibility of 3D reconstruction and consider time constraints, it is recommended to set the number of candidate views to 5.

Of course, these trends are generally applicable. Therefore, in different datasets or applications, the reasonable selection of the number of views needs to be determined through experiments to ensure the accuracy of this method in practical use. 
\begin{table}[ht]
	\newcommand{\tabincell}[2]{\begin{tabular}{@{}#1@{}}#2\end{tabular}}
	\centering
	\caption{Comparison of Methods on the Hpatch Image Dataset}
	\setlength{\tabcolsep}{0.5pt}
	\begin{tabular}{ccccc}
	\toprule
	\tabincell{c}{Parameter} & \multicolumn{2}{c}{100-image dataset} & \multicolumn{2}{c}{500-image dataset}  \\
	\cmidrule(r){2-3} \cmidrule(lr){4-5} 
	& \tabincell{c}{Reprojection \\ error}  &  Time   & \tabincell{c}{Reprojection \\ error}  &  Time \\
	1	 &$\times$	 &$\times$	 &$\times$	 &$\times$\\
2	 &2.0438 	 &3.700 &$\times$	 &$\times$\\
3	 &1.9121 	 &3.455	 &926.9672	 &20.705\\
4	 &1.3648 	 &3.884	 &899.9288	 &21.370\\
5	 &1.3469 	 &4.297	 &576.5797	 &21.774\\
6	 &1.2089 	 &4.580	 &877.1375	 &23.360\\
7	 &1.2265 	 &5.272	 &718.0292	 &23.851\\
8	 &1.2427 	 &5.229	 &1395.629	 &28.590\\
9	 &1.0606 	 &5.296	 &443.0745	 &28.017\\
10	 &1.1300 	 &5.700	 &667.6387	 &30.862\\
	\bottomrule
	\end{tabular}
	\end{table}
In addition, when applying our view selection method in the COLMAP program, as our method requires baseline and 3D point coordinates to calculate error resistance values for triangulation, which are unknown at the beginning, it is necessary to perform an initial coarse reconstruction using COLMAP with an exhaustive method. This allows rapid acquisition of camera baselines and corresponding 3D points. To expedite the coarse reconstruction, default parameter modifications are made to the time-consuming bundle adjustment step. This is because, in the SfM process, for the reconstruction of each view, after estimating camera poses and triangulation, a bundle adjustment is performed to optimize the reprojection error function. This optimization involves a relatively large number of iterations compared to other steps, leading to longer processing times. Therefore, the iteration parameters for bundle adjustment after camera pose estimation and triangulation in the COLMAP program, namely "max$\_$num$\_$iterations" and "max$\_$refinements," are set to 1 to minimize the time required for coarse reconstruction.

Simultaneously, as our view selection method is based on the triangulation error resistance values of baselines, the optimal baselines under this criterion correspond to two views with noticeable variations. If a method more suitable for wide-baseline feature extraction and matching is not used, fewer corresponding feature point pairs are retained after filtering. To ensure that all views can participate in 3D reconstruction in the COLMAP program, the conditions for the feature matching step are relaxed. Specifically, the minimum inlier count is changed to 15, and the minimum inlier ratio is set to 0.1.

\subsection{Experimental Comparison}
(1) 3D Reconstruction Accuracy Experiment

Based on the COLMAP program, the view selection method proposed in this study is integrated and compared with the most accurate exhaustive method. The comparison focuses on the average reprojection error of 3D points and the absolute trajectory error of camera poses.

The experimental results on the TUM dataset are illustrated in Figures 3, 4, and 5. Figure 3 shows that, using the average reprojection error as the evaluation metric, the results of this method are significantly lower than those of the exhaustive method, indicating that the precision of 3D point reconstruction supported by this view selection method is superior. Figures 4 and 5 present box plots for the absolute trajectory error experiments on the "freiburg1$\_$teddy" and "freiburg1$\_$xyz" sequences, respectively. The x-axis represents the sampling interval of image sets, where a larger interval indicates a greater camera baseline. The rectangular region of the box plot represents the range where 50$\%$ of the data values fall, the line inside the box indicates the data's average, and the short lines at the top and bottom denote the maximum and minimum values.

In Figure 5, due to the significant fluctuation in error over a wide range, the box corresponding to image set intervals of 4 and 6 exceeds the numerical range and cannot be fully displayed. However, this does not affect the comparison of average values. Therefore, to present more data comprehensively, the vertical axis range is not expanded here. Figures 4 and 5 demonstrate that the average absolute trajectory error of this method is lower than that of the exhaustive method for image sets with different baselines. The 50\% data range under 7 different baselines is generally lower for this method compared to the exhaustive method, indicating that the majority of errors calculated by this method for all camera poses are lower than those of the exhaustive method, resulting in superior outcomes. The only drawback is that, under most baselines, the maximum error for this method is higher than that for the exhaustive method. According to calculations, the average reprojection error is reduced by 28.84\%, and the average absolute trajectory error is reduced by 5.29\% on the TUM dataset using this method. Therefore, this view selection method maintains better reconstruction results on image sets with different intervals, i.e., different baselines.

\begin{figure}[H]	
	\centering
	\begin{subfigure}{0.45\linewidth}
		\centering
		\includegraphics[width=0.99\linewidth]{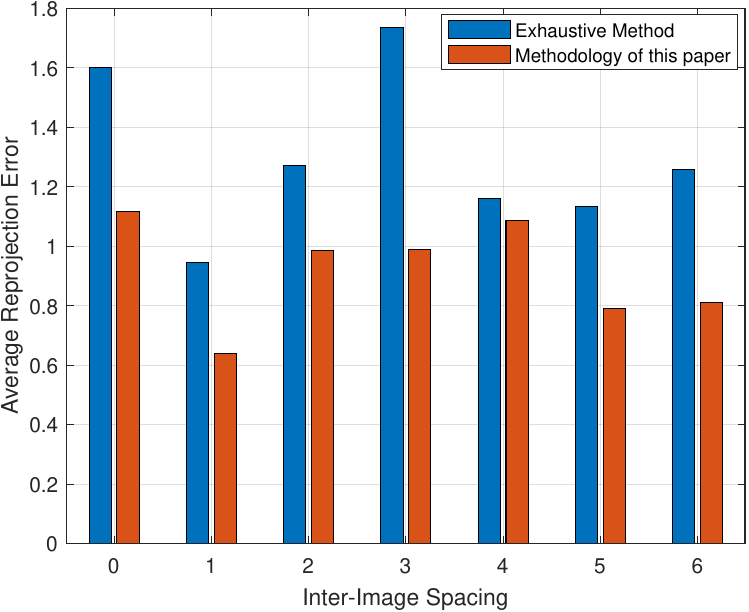}
		\caption{"freiburg1\_teddy" sequence}
		\label{Fig41}%
	\end{subfigure}
	\centering
	\begin{subfigure}{0.45\linewidth}
		\centering
		\includegraphics[width=0.99\linewidth]{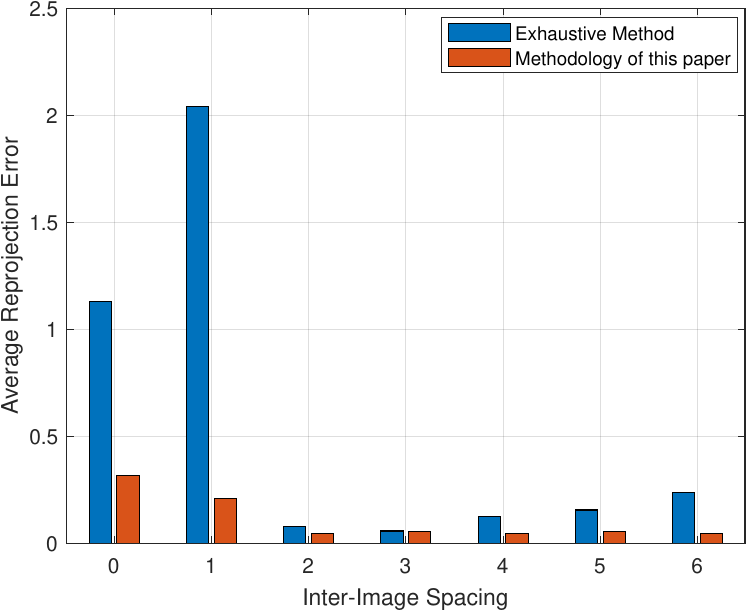}
		\caption{"freiburg1\_xyz" sequence}
		\label{Fig42}
	\end{subfigure}
	\caption{Reprojection error results on the TUM dataset.}
\end{figure}

\begin{figure}[H]	
	\centering
	\begin{subfigure}{0.45\linewidth}
		\centering
		\includegraphics[width=0.99\linewidth]{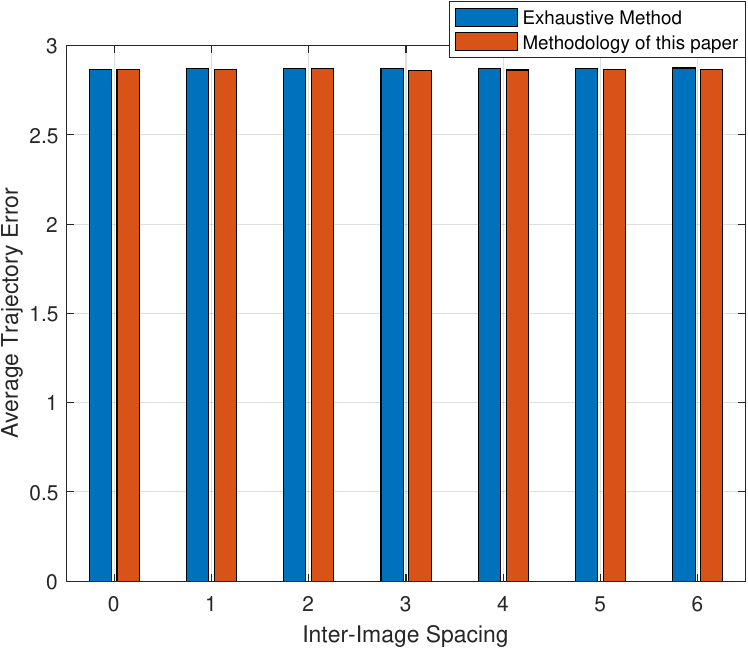}
		\caption{"freiburg1\_teddy" sequence}
		\label{Fig51}%
	\end{subfigure}
	\centering
	\begin{subfigure}{0.45\linewidth}
		\centering
		\includegraphics[width=0.99\linewidth]{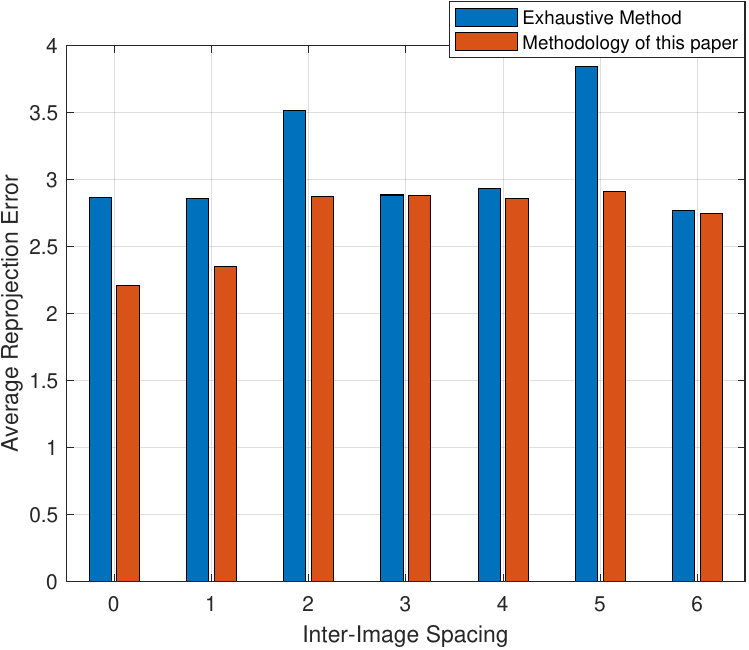}
		\caption{"freiburg1\_xyz" sequence}
		\label{Fig52}
	\end{subfigure}
	\caption{Absolute trajectory error results based on the TUM dataset.}
\end{figure}
The experimental results on the DTU dataset are presented in Table 2. Among the 20 image sets, the proposed method outperforms the exhaustive method in terms of both absolute trajectory error and average reprojection error in 15 image sets. This indicates that the proposed method demonstrates superiority over the exhaustive method in 75\% of the scenes in the dataset. The average absolute trajectory error for the 20 image sets, as calculated from Table 2, is 2.1247 meters for the proposed method and 2.2330 meters for the exhaustive method, resulting in an improvement of 4.85\%. Similarly, the average reprojection error for the 20 image sets is 5.7048 pixels for the proposed method and 8.1426 pixels for the exhaustive method, showing an improvement of 29.95

\begin{table}[ht]
	\newcommand{\tabincell}[2]{\begin{tabular}{@{}#1@{}}#2\end{tabular}}
	\centering
	\caption{Comparison of Methods on the Hpatch Image Dataset}
	\setlength{\tabcolsep}{0.5pt}
	\begin{tabular}{ccccc}
	\toprule
	\tabincell{c}{Parameter} & \multicolumn{2}{c}{Absolute trajectory error} & \multicolumn{2}{c}{Average reprojection error}  \\
	\cmidrule(r){2-3} \cmidrule(lr){4-5} 
	& \tabincell{c}{Exhaustive \\ method}  &  \tabincell{c}{This paper's\\ method}   & \tabincell{c}{Exhaustive \\ method}  &  \tabincell{c}{This paper's\\ method}  \\
Scan1	&2.0949	&2.2768	&6.1695	&5.7646\\
Scan2	&2.3301	&2.1307	&3.9312	&2.6900\\
Scan3	&2.2511	&2.1129	&10.1503	&5.7750\\
Scan4	&2.1434	&2.0356	&5.0087	&4.5988\\
Scan5	&2.0979	&2.1022	&4.6478	&4.3992\\
Scan6	&2.2379	&2.2579	&6.0278	&5.2664\\
Scan7	&2.3452	&2.1306	&4.4385	&4.1486\\
Scan8	&2.5486	&2.2323	&40.1290	&10.8076\\
Scan9	&2.5158	&2.1708	&16.1923	&8.5981\\
Scan10	&2.3365	&2.0223	&4.6768	&5.6369\\
Scan11	&2.2792	&2.0290	&15.6755	&8.0161\\
Scan12	&2.2032	&2.0151	&5.9162	&2.0157\\
Scan13	&2.1459	&2.0239	&4.1090	&4.5624\\
Scan14	&2.1094	&2.1425	&4.9605	&3.5521\\
Scan15	&2.2572	&2.0251	&5.2396	&5.1048\\
Scan16	&2.0666	&2.0508	&3.4308	&3.6828\\
Scan17	&2.2256	&2.1738	&4.3933	&4.9826\\
Scan18	&2.1147	&2.1572	&6.2432	&5.7513\\
Scan19	&2.1887	&2.0115	&5.6152	&6.2152\\
Scan20	&2.1676	&2.1946	&5.8968	&5.3569\\
	\bottomrule
	\end{tabular}
	\end{table}
(2) Runtime Comparison

A comparison of the runtime between the COLMAP program with the proposed method and the exhaustive method is conducted under the following experimental conditions:

Hardware Environment:CPU: Intel(R) Core(TM) i7-10875H Memory: 16GB GPU: NVIDIA GeForce RTX 2060 6GB

Software Environment: IDE: PyCharm Professional 2020.1 Dependencies: OpenCV-Python library version 4.5.3, NumPy library version 1.19.2 COLMAP Program: Officially compiled program run via the command line

Under these conditions, the specific runtime results are presented in Table 3. The runtime for the proposed method includes the time for coarse reconstruction, view selection, and feature matching. In contrast, the exhaustive method's view selection primarily involves the time for feature matching since it matches features between all pairs of views. Although the proposed method adds extra time for coarse reconstruction, it reduces the number of iterations in the Bundle Adjustment (BA) stage during the reconstruction phase, resulting in no significant increase in overall runtime.

\begin{table*}[!hptb]
	\centering
	\caption{Running time}
	\setlength{\tabcolsep}{0.3pt}
	\begin{tabular}{l *{4}{S[table-format=2.4]}}
	  \toprule
	  & & {View selection phase (min)} & {Reconstruction time (min)} & {Total time (min)} \\
	  \midrule
	  \multirow{2}{*}{freiburg1\_teddy} & {Exhaustive method} & 2.088 & 10.369 & 12.457 \\
	  & {This paper's method} & {4.419} & 4.097 & 8.695 \\ 
	  \hline
	  \multirow{2}{*}{freiburg1\_xyz} & {Exhaustive method} & 0.144 & 9.479 & 9.623 \\
	  & {This paper's method} & {1.291} & 4.306 & 5.644 \\
	  \hline
	  \multirow{2}{*}{DTU} & {Exhaustive method} & 0.032 & 0.366 & 0.398 \\
	  & {This paper's method} & {0.254} & 0.320 & 0.594 \\ 
	  \bottomrule
	\end{tabular}
	\label{tab3}
\end{table*}

  (3) Method Effect Demonstration

To showcase the 3D reconstruction results achieved through the proposed view selection and its application in COLMAP, a set of image sequences from the DTU dataset is chosen for presentation. The results are illustrated in Figures 4-6.

In Figures 4-6(a), each solid square represents a camera, and the lines between cameras depict the candidate relationships for the next view. In Figures 4-6(b), each solid square represents a camera, and the direction of the camera points towards the reconstructed point cloud of the 3D object. 

\begin{figure}[H]	
	\centering
	\begin{subfigure}{0.45\linewidth}
		\centering
		\includegraphics[width=0.99\linewidth]{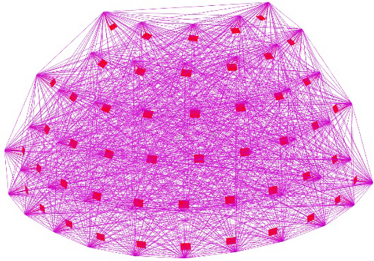}
		\caption{View Selection Results}
		\label{Fig61}%
	\end{subfigure}
	\centering
	\begin{subfigure}{0.45\linewidth}
		\centering
		\includegraphics[width=0.99\linewidth]{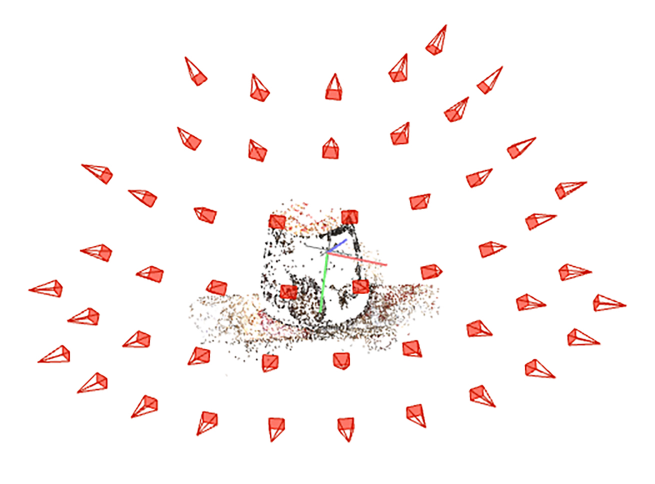}
		\caption{3D Reconstruction Results}
		\label{Fig62}
	\end{subfigure}
	\caption{Effect Display Diagram.}
\end{figure}

\section{Conclusion}
Starting from the principles of spatial imaging, this paper analyzes the impact mechanism of camera baselines on the robustness of Structure-from-Motion (SfM) triangulation. A view selection method based on camera baselines is proposed, which calculates triangulation resistance values under different camera baselines to support a more rational selection of candidate views. Simultaneously, based on the basic process of selecting the next view for reconstruction in SfM, the candidate view set is examined and completed to ensure the integrity of 3D reconstruction. Finally, this view selection method is integrated into the COLMAP program. Using the reprojection error of 3D point clouds and the absolute trajectory error of cameras as indicators, experiments validate and evaluate the advantages of this method in supporting 3D reconstruction compared to the widely recognized exhaustive method on different datasets. 





\end{document}